\documentclass[letterpaper]{article} 
\usepackage{aaai2027} 
\usepackage[hyphens]{url} 
\usepackage{graphicx} 
\usepackage{natbib} 
\usepackage{caption} 
\usepackage{algorithm}
\usepackage{algorithmic}
\usepackage{amsfonts}
\usepackage{amsmath}
\usepackage{booktabs}

\usepackage{newfloat}
\usepackage{listings}
\DeclareCaptionStyle{ruled}{labelfont=normalfont,labelsep=colon,strut=off} 
\floatstyle{ruled}
\newfloat{listing}{tb}{lst}{}
\floatname{listing}{Listing}

\usepackage{booktabs}

\title{OfficeTwin: Realizing 3D Reconstruction of Redundant Scenes through Partial Viewpoint Perception}
\author{
 Quanyun Wu {\rm 1},
 Kyle Gao{\rm 2},
 Daisy Li{\rm 1},
 Kshitij Goyal {\rm 1},
 Dayou Mao {\rm 1},
 David A. Clausi{\rm 1},
 Jonathan Li {\rm 1},
 Yuhao Chen{\rm 1}
}
\affiliations{
 \textsuperscript{\rm 1}University of Waterloo\\

 \textsuperscript{\rm 2}Aalto University\\
}

\begin{document}

\maketitle

 \begin{abstract}
Feed-forward 3D reconstruction turns an uncalibrated monocular stream into a dense point cloud in a single inference pass, orders of magnitude faster than classical structure-from-motion (e.g., COLMAP), and reliable on the weakly textured, repetitive surfaces where keypoint matching fails outright. However, raw point clouds lack the object identity, physical properties, and watertight volume required for robot learning. This gap widens under robot-native capture. A camera mounted on a robot constrained to a few degrees of freedom sees each object only partially, and the repetitive, weakly textured structure of an office makes distinct workstations nearly indistinguishable, leaving redundant and partial observations that downstream reconstruction cannot resolve. Existing image-to-3D generation methods struggle with these conditions, hallucinating geometry under incomplete views and confusing repeated instances. We present OfficeTwin, a robot-native framework that converts these incomplete observations into simulation-ready digital twins. OfficeTwin first imposes metric and structural constraints, then exploits the office's structural redundancy by registering repeated cubicles into a shared canonical frame, where complementary observations recover geometry occluded in any single view without hallucination. On a real office scan, OfficeTwin recovers every cubicle as an individually watertight, metric solid, outperforming feed-forward reconstruction and generative baselines that fail to produce simulation-ready assets.

\end{abstract}


\section{Introduction}

Feed-forward 3D reconstruction now recovers dense 3D geometry from a monocular RGB stream in a single forward pass, without the slow per-scene optimization of classical multi-view stereo or the per-scene fitting of radiance fields~\cite{deng2025vggt, wang2024dust3r}. For Embodied AI, Sim-to-Real transfer requires simulated environments to faithfully reproduce absolute metric scale and watertight geometry that govern obstacle avoidance and contact-rich manipulation. To scale to arbitrary environments, twins should be generated directly from the robot's onboard RGB camera along traversable trajectories, avoiding costly operator scanning. Yet these reconstructions remain unusable as simulation assets: they lack structure and watertight geometry.
\begin{figure*} [!t]
 \centering
 \includegraphics[width=1\linewidth]{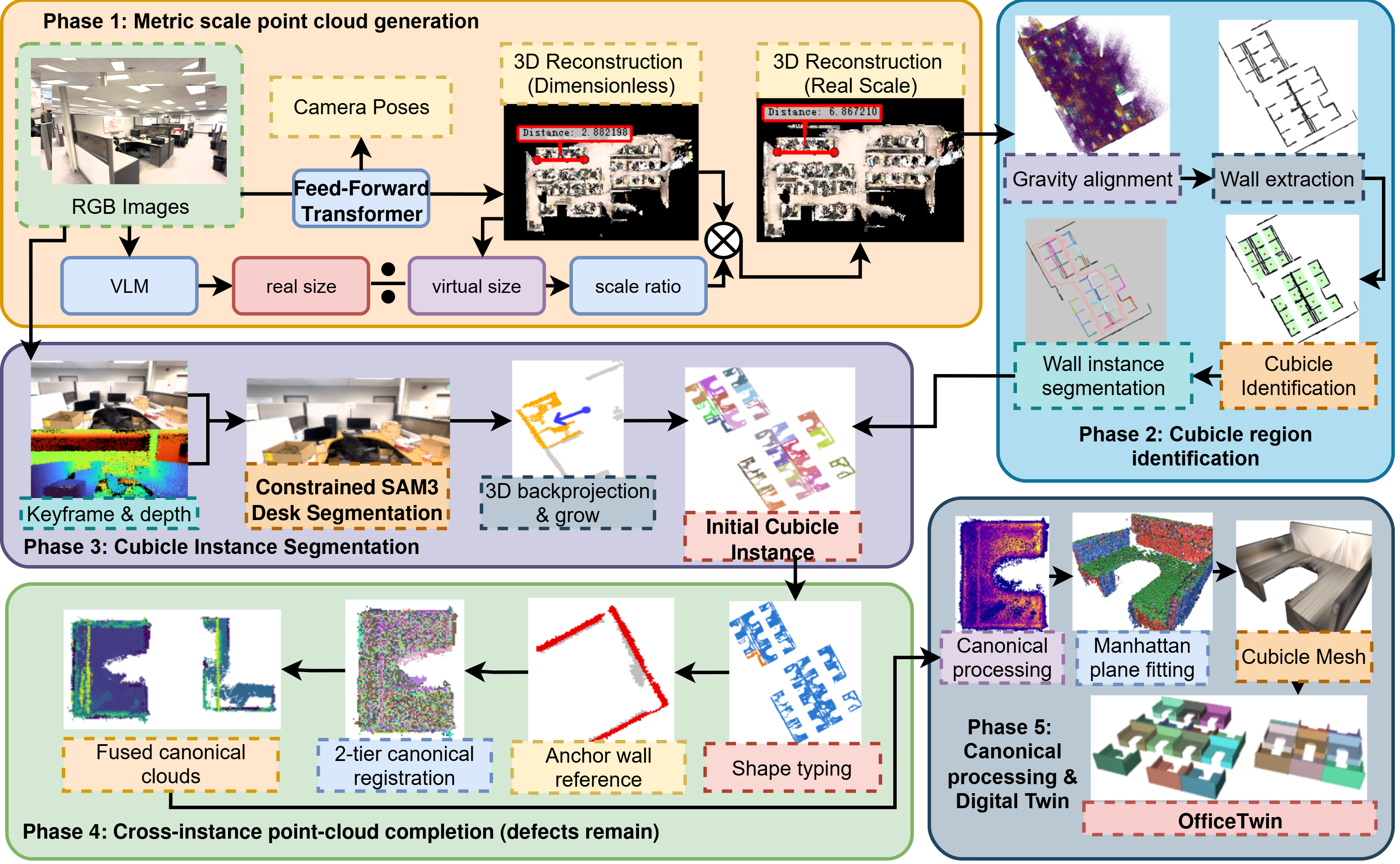}
 \caption{Overview of the OfficeTwin pipeline, converting RGB frames of a redundant office into a watertight, metrically scaled Manhattan digital twin in five phases: (1)~VLM-based metric scaling; (2)~gravity-aligned structural parsing and cubicle identification; (3)~wall-bounded 3D SAM3 instance growth; (4)~canonical cross-instance registration for occlusion completion; and (5)~consensus denoising and Manhattan solid fitting.}
 \label{fig:pipeline}
\end{figure*}
Converting raw point clouds into structured, watertight simulation assets is difficult, particularly for robot-captured office imagery, for two reasons. First, body-mounted cameras constrained to narrow aisles suffer from systematic occlusions, observing every workstation only partially (\emph{partial view}). Second, an office is filled with near-identical, weakly textured cubicles (\emph{redundant structure}).

As shown in Fig.~\ref{fig:baselines}, existing methods that try to turn these partial, redundant observations into a clean, per-object 3D model fail under these conditions. Single-image generators such as SAM3D~\cite{chen2025sam}, Hunyuan3D-2~\cite{zhao2025hunyuan3d}, and TRELLIS~\cite{xiang2025structured} hallucinate or merge neighbouring instances under partial and redundant views. Feed-forward reconstruction provides dense geometry but remains scale-free and topologically incomplete~\cite{xiong2026vggt, im2026prism}, while learned completion lacks explicit structural constraints~\cite{wang2026dehallu3d, liu2026had}. These failures share a root cause: the initial reconstructions are scale-free, structure-blind, and incomplete. Repairing them requires jointly: (i)~recovering absolute metric scale; (ii)~achieving instance-correct segmentation despite destabilizing redundancies; and (iii)~completing and closing geometry left open by partial views, without hallucination or pose-drift ghosting. Our central observation is that the office's redundancy, the very property that defeats prior methods, makes the third and hardest sub-problem solvable. Every cubicle recurs many times. A region occluded in one instance is almost always observed in another, so repeated instances can complete one another rather than being invented.

As shown in Fig.~\ref{fig:pipeline}, we propose OfficeTwin, a robot-native pipeline that turns these obstacles into structure and addresses the three sub-problems in turn. To recover scale, \emph{semantic metric anchoring} converts a real-world dimension read by a VLM into a single global metric factor. To make segmentation instance-correct, \emph{structure-aware grounding} reprojects the reconstructed 3D wall topology onto each view as a hard constraint that restricts the foundation model to one cubicle at a time. To complete missing geometry, \emph{redundancy-driven completion} registers repeated cubicle instances into a shared canonical frame, allowing complementary views to fill occlusions. Finally, it removes residual ghosting via multi-view consensus and fits watertight Manhattan solids. Unlike pipelines that isolate semantic segmentation from geometric meshing~\cite{hane2016dense, feng2024terrainmesh}, our components share the recovered scene structure throughout the entire process.

In summary, our contributions are:
\begin{itemize}
 \item \textbf{Semantic metric anchoring}: a VLM-driven mechanism that recovers absolute metric scale for an otherwise scale-free feed-forward reconstruction.
 \item \textbf{Structure-aware foundation grounding}: reprojecting the reconstructed 3D wall topology as hard constraints that make a 2D foundation segmentation model instance-correct in repetitive layouts, reducing the mask spillover and over-segmentation these models otherwise exhibit.
 \item \textbf{Redundancy-driven completion}: a canonical cross-instance registration that exploits architectural repetition to fill partial-view occlusions, followed by multi-view consensus denoising and Manhattan solid fitting for watertight, metrically scaled geometry.
\end{itemize}

\section{Related Work}
\textbf{Feed-Forward Foundation Models and Scalability.} Recent feed-forward transformers estimate visual geometry without slow per-scene optimization~\cite{campos2021orb, keetha2024splatam}, regressing structure directly from RGB inputs to bypass traditional feature-matching failures in textureless scenes. While architectures like VGGT~\cite{wang2025vggt} and $\pi^3$~\cite{wang2025pi} resolve camera poses and local geometry, their scaling remains constrained by quadratic memory growth. Long-sequence extensions like VGGT-Long~\cite{deng2025vggt} and Pi-Long~\cite{deng2025pilong} leverage chunk-based reconstruction and $\mathrm{Sim}(3)$ alignment to mitigate this limitation, yet these pure geometry-driven approaches cannot resolve missing surfaces or ghosting artifacts caused by partial observations. OfficeTwin adopts the scalable chunking strategy of Pi-Long~\cite{deng2025pilong} but introduces voxel consensus and structural reasoning to derive simulation-ready reconstructions.

\begin{figure*} [!t]
\centering
\includegraphics[width=\textwidth]{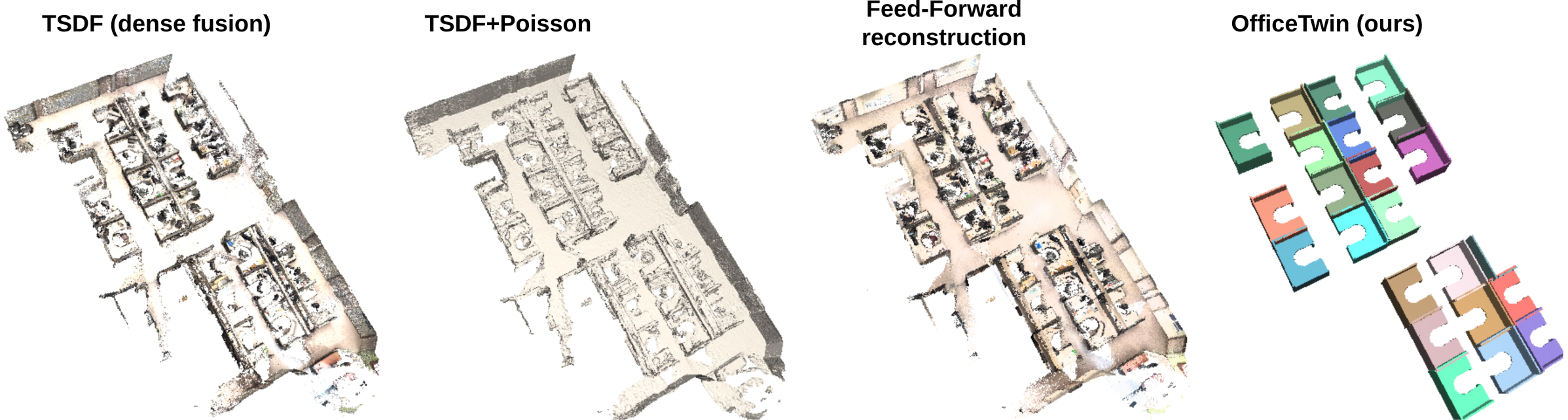}
\caption{Full-scene reconstruction (ceiling removed). From left: TSDF fusion yields an inseparable hole-riddled mesh and structureless Poisson surface; Pi-Long/VGGT-Long yields a scale-ambiguous cloud; OfficeTwin recovers $23$ individual, watertight metric solids instead of one fused mass.}
\label{fig:comparison}
\end{figure*}

\textbf{Visual Foundation Models and 3D Segmentation.} Segment-anything architectures have advanced zero-shot instance segmentation. SAM3~\cite{carion2025sam} and its multi-modal 3D counterpart SAM3D~\cite{chen2025sam} extend this to open-vocabulary understanding, while several works lift 2D models into 3D via feature projection~\cite{kerr2023lerf}. Applied directly to indoor imagery, however, these models lack structural constraints and struggle to separate overlapping, repetitive instances (e.g., adjacent desks), resulting in mask spillover. OfficeTwin instead grounds segmentation in scene geometry, projecting reconstructed wall topology into each view as an exclusion region for SAM3.

\textbf{Occlusion Completion.} Unobserved scan regions are traditionally addressed via three paradigms. Generative and diffusion models synthesize plausible geometry~\cite{liu2023zero, wu2024reconfusion} but frequently introduce physical layout violations detrimental to Sim-to-Real applications. Learned completion infers missing surfaces using self-supervised external datasets or local structural patch priors~\cite{dai2020sg, rao2022patchcomplete}, which remain unanchored to the specific scene context and prone to hallucinating inaccurate details. Structural fitting approaches optimize polygonal surfaces or structured layout topologies~\cite{nan2017polyfit, chen2022heat} but assume a globally complete input scan. Conversely, OfficeTwin requires neither external priors nor a complete scan: it completes each workstation from the complementary observations of its repeated instances, keeping only surfaces that several instances corroborate. Completion thus stays anchored to observed geometry, yielding a watertight solid per instance
 rather than a plausible surface.

\textbf{Digital Twins for Sim-to-Real Robotics.} High-fidelity Digital Twins are a key enabler for scaling robotic learning~\cite{szot2021habitat}, and Sim-to-Real paradigms such as mapless navigation require training environments that match real-world physics~\cite{yang2025sru}. By construction, feed-forward models such as $\pi^3$~\cite{wang2025pi} and pipelines such as Pi-Long~\cite{deng2025pilong} output geometry only up to an arbitrary Sim(3) transformation; this scale ambiguity biases the spatial representations a Reinforcement Learning (RL) agent learns, for instance distorting grasp distances and obstacle clearances. OfficeTwin resolves this via a VLM-driven scaling stage that reads real-world dimensions from visual landmarks and, via resolution-aligned frustum culling, anchors the output to absolute metric scale.

\section{Methodology}

The OfficeTwin pipeline takes as input a scale-ambiguous point cloud $\mathcal{P}_{raw}$ reconstructed from a monocular RGB sequence by a feed-forward reconstruction model, and converts it into a metrically scaled, watertight Manhattan mesh $\mathcal{M}$ through the five phases of Fig.~\ref{fig:pipeline}. The pipeline successively recovers metric scale, identifies cubicle topology, segments individual workstations, registers repeated cubicles into a shared canonical frame, removes residual ghosting by multi-view consensus, and finally fits watertight Manhattan solids. Throughout, repeated cubicle layouts serve as internal supervision, allowing incomplete geometry to be recovered from complementary observations rather than learned priors.

\subsection{Metric Scale Recovery}

Feed-forward reconstruction is defined only up to a similarity ($\mathrm{Sim}(3)$) transformation. We therefore recover the missing metric scale from a reference object with semantically identifiable dimensions. A Vision-Language Model selects a fronto-parallel object in a reference frame $I_{ref}$ and returns its real-world width and height $(w,h)$ together with a 2D bounding box $\mathcal{B}$. Since the VLM operates on the full-resolution image while the reconstruction carries the intrinsics of the down-sampled stream, $\mathbf{K}$ is rescaled to the VLM resolution before any projection. We then isolate the object's 3D points by frustum culling followed by depth gating,
\begin{equation}
\begin{split}
\mathcal{P}_{obj}=\Big\{\,\mathbf{p}\in\mathcal{P}\ \Big|\ & \pi_{\mathbf{K}}\!\big(\mathbf{T}_{ref}\,\mathbf{p}\big)\in\mathcal{B}\ \wedge \\
& z\big(\mathbf{T}_{ref}\,\mathbf{p}\big)\in[d_{lo},d_{hi}]\,\Big\},
\end{split}
\end{equation}
where $\mathbf{T}_{ref}$ is the world-to-camera transform of $I_{ref}$, $\pi_{\mathbf{K}}(\mathbf{x})=(f_x x/z+c_x,\,f_y y/z+c_y)$ is the pinhole projection, and points with $z\le z_{near}$ are discarded.The depth gate $[d_{lo},d_{hi}]$ is the maximal contiguous interval around the modal bin of the depth histogram within which the bin counts remain above half the modal count; this filter removes background surfaces inevitably trapped inside $\mathcal{B}$. A statistical outlier filter is applied before measuring the extent, since axis-aligned bounds are determined by extrema. The global scale factor is computed as
\begin{equation}
s=\frac{\sqrt{w^2+h^2}}
{\left\|\max(\mathcal{P}_{obj})-\min(\mathcal{P}_{obj})\right\|_2},
\end{equation}
and the reconstructed scene and camera translations are rescaled by $s$. Requiring a fronto-parallel reference object keeps the numerator, a diagonal of the object's visible face, comparable to the denominator, the diagonal of its 3D axis-aligned bounds.

\subsection{Gravity-Aligned Structural Parsing}

As shown in Fig.~\ref{fig:topology}, repeated cubicle layouts make instance clustering unreliable, as neighbouring workstations share similar appearance and remain spatially adjacent. OfficeTwin therefore recovers structural topology, using gravity-aligned Manhattan walls as stable geometric boundaries to identify individual cubicle regions trajectory-independently. We estimate the gravity axis $\hat{\mathbf{g}}$ as the normal of the dominant horizontal plane using RANSAC on a down-sampled cloud, and rotate the scene by $\mathbf{R}_g$ such that $\mathbf{R}_g\hat{\mathbf{g}}=\mathbf{e}_y$. On the gravity-aligned cloud, we retain near-vertical points ($|\mathbf{n}_p\!\cdot\!\mathbf{e}_y|<\tau_v$) and estimate two orthogonal Manhattan directions from a length-weighted histogram of horizontal normals. For a wall orientation with normal $\hat{\mathbf{n}}(\theta)=(\cos\theta,\sin\theta)$, each point is represented by its Hesse offset
\begin{equation}
\rho_p=\mathbf{p}^{xz}\!\cdot\hat{\mathbf{n}}(\theta).
\end{equation}
Gap splitting over $\{\rho_p\}$ recovers individual wall planes. Planes satisfying a joint length-density criterion are classified as partitions $\mathcal{W}_p$ or back walls $\mathcal{W}_b$, separating architectural boundaries from isolated tall clutter. We then localize individual workstations from the desk regions enclosed by these structural walls. Desk-height points are rasterized into a top-down occupancy grid, morphologically closed, and segmented by the dilated wall set $\mathcal{W}=\mathcal{W}_p\cup\mathcal{W}_b$ into connected components $\{R_k\}$. A component is accepted as a cubicle when its metric area lies in $[A_{\min},A_{\max}]$ and its boundary contacts both wall classes:
\begin{equation}
\begin{aligned}
A_{\min}\!\le\!|R_k|\,\delta^{2}\!\le\!A_{\max}\ \wedge\ 
\partial R_k\!\cap\!\mathcal{W}_p\!\neq\!\emptyset\\
\wedge\ 
\partial R_k\!\cap\!\mathcal{W}_b\!\neq\!\emptyset .
\end{aligned}
\end{equation}
Each accepted region is completed into its wall-enclosed rectangle by intersecting its minimum-area rectangle with wall-free space, producing a closed footprint without trajectory-dependent thresholds. Finally, from each cubicle center $c_k$, we cast $N_r=1440$ line-of-sight rays to assign wall points only when they match the first visible surface within a strict visibility thickness $\tau_{vis}$ of the ray hit. This visibility assignment accurately separates shared partitions between adjacent cubicles, preventing nearest-distance ambiguity.
\begin{figure*}[!t]
\centering
\includegraphics[width=\textwidth]{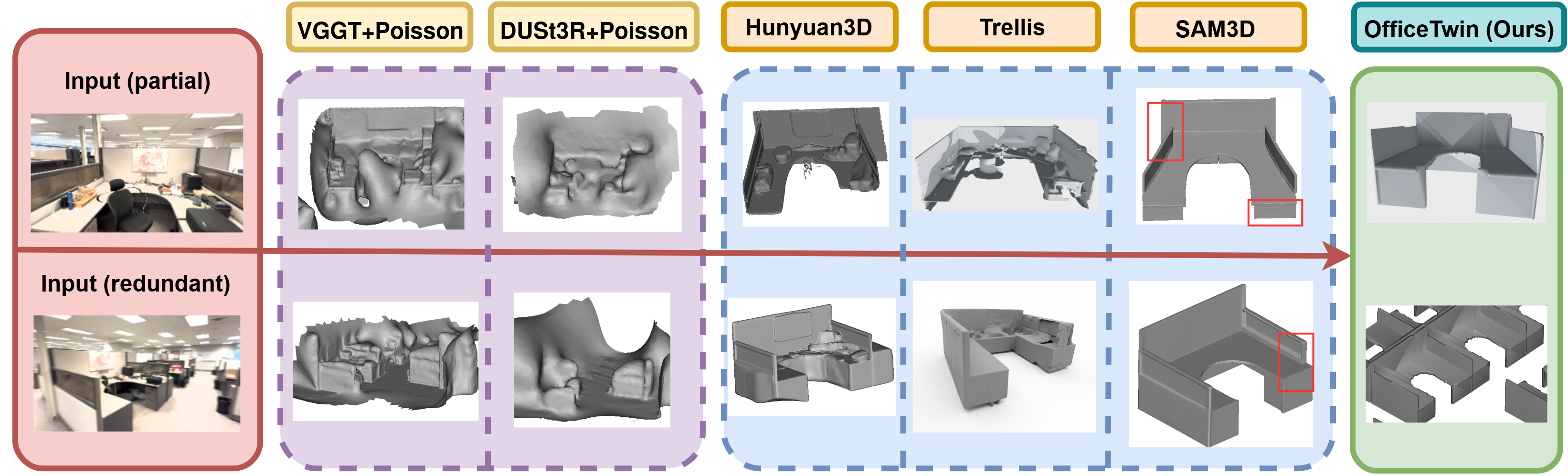}
\caption{Single-cubicle reconstruction from a partial view (top) and a redundant multi-cubicle frame (bottom). Left to right: feed-forward reconstruction with Poisson surfacing (purple; ceiling removed), single-image generative reconstruction (blue), and OfficeTwin (green). Red boxes mark the SAM3D failures analysed in Sec.~\ref{subsec:partial}}
\label{fig:baselines}
\end{figure*}

\subsection{Geometry-Grounded Instance Segmentation}

Geometry-grounded 2D segmentation: Each cubicle is assigned to the keyframe $f_k^{\star}$ that provides the best observation of its center $c_k$, selected by visibility and viewing distance. Projecting its wall points $\mathcal{W}_k$ into $f_k^{\star}$ and keeping depth-consistent pixels gives an occlusion-aware wall mask forbidden to the foundation model,
\begin{equation}
\begin{aligned}
\mathcal{M}_{wall} =\Big\{ \pi_{f_k^{\star}}(p)\ \Big|\ 
p\in\mathcal{W}_k,\\
z(p)\in[\mathcal{D}(u,v)-\delta_1,\,
\mathcal{D}(u,v)+\delta_2]
\Big\},
\end{aligned}
\end{equation}
where $\mathcal{D}$ is the rendered depth of $f_k^{\star}$. From the SAM3 desk candidates $\{\mathcal{M}\}$, we select the largest mask that strictly obeys the wall boundaries, suppressing spillover onto identical neighbours:
\begin{equation}
\mathcal{M}^{\star}=\arg\max_{\mathcal{M}}|\mathcal{M}|
\quad\text{s.t.}\quad
\frac{|\mathcal{M}\cap\mathcal{M}_{wall}|}{|\mathcal{M}|}\le\tau_{climb}.
\end{equation}

Wall-bounded 3D growth: $\mathcal{M}^{\star}$ is back-projected to desk-height seed points, whose mean normal $\bar{\mathbf{n}}$ and height $\bar{y}$ define a local desk prior. A candidate $q$ is added to instance $\mathcal{I}_k$ when it is normal-consistent with the desk, lies on the desk plane, and remains inside the cubicle's wall bounding box:
\begin{equation}
|\mathbf{n}_q\!\cdot\!\bar{\mathbf{n}}|\ge\tau_n,\quad
|y_q-\bar{y}|\le\tau_y,\quad
(x_q,z_q)\in\mathrm{bbox}(\mathcal{W}_k).
\end{equation}
The wall cage recovers regions absent from SAM3 observations while preventing instance spillover caused by the near-identical appearance of neighbouring cubicles. The instance labels are thus defined in 3D, with depth entering only through the 2D mask $\mathcal{M}_{wall}$.

\subsection{Cross-Instance Canonical Registration}

Different instances reveal complementary regions under partial views. OfficeTwin leverages this spatial redundancy, registering same-shape instances into a shared canonical frame to recover occluded geometry. \textbf{Manhattan Frame Estimation:} Alignment presupposes a shared reference orientation. For each cubicle $k$, the raw yaw $\alpha_k \in [0, \tfrac{\pi}{2})$ about the gravity axis $\mathbf{e}_y$ is extracted from the length-weighted wall normal histogram. To resolve the $\pi/2$-periodicity ambiguity of Manhattan layouts where raw averaging fails, we project the angles onto a unit circle by quadrupling the phase. The global Manhattan yaw $\bar{\alpha}$ is determined via the circular mean:
\begin{equation}
    \bar{\alpha}=\tfrac{1}{4}\arg\Big(\textstyle\sum_{k}e^{\,i\,4\alpha_k}\Big)
\end{equation}
Rotating the scene by $-\bar{\alpha}$ ensures all cubicle footprints are axis-aligned. \textbf{Shape Typing \& Pose Faces:} For each footprint edge $e$, we compute the wall-coverage ratio $\mathrm{cov}(e)=\tfrac{1}{N_b}\sum_{b}\mathbb{I}(h_b(e)>0)$, where $N_b$ is the total length bins and $h_b(e)$ represents the wall points falling into bin $b$. Instances are classified by topology: three covered edges indicate a U-shape, while two adjacent covered edges indicate an L-shape. The uncovered edge determines the instance orientation $\psi_k$. To prevent overhead cabinets from introducing point-to-plane ICP ambiguities, we isolate the outermost plane of each covered edge as the pose face $\mathcal{F}_k$ using a $1/99$-percentile depth offset gate $\tau_f$. \textbf{Two-Tier Alignment:} Same-type instances are registered to a selected anchor cloud. A normalizing transform $\mathbf{T}_0(\psi_k,c_k)$ first canonicalizes the orientation $\psi_k$ and centers the instance $c_k$. A coarse 2D alignment $\mathbf{T}_1$ then optimizes an edge-skeleton cost combining the anchor's distance transform $\mathbf{DT}_a$ and template correlation $\mathrm{TM}$:
\begin{equation}
    \mathbf{T}_1=\arg\min_{T}\ \frac{\frac{1}{\vert{}\mathcal{E}_k\vert{}}\sum_{x\in\mathcal{E}_k}\mathbf{DT}_a(Tx)}{\mathrm{TM}(\mathcal{E}_a,\,T\mathcal{E}_k)+\epsilon}
\end{equation}
where $\mathcal{E}_a$ and $\mathcal{E}_k$ denote the anchor and template edge skeletons, respectively. Finally, a Tier-2 point-to-plane ICP $\mathbf{T}_2$ refines the alignment strictly on the pose faces:
\begin{equation}
    \mathbf{T}_2=\arg\min_{T}\ \sum_{i}\big((T\mathbf{f}_i-\mathbf{a}_{j(i)})\!\cdot\!\mathbf{n}_{j(i)}\big)^{2}
\end{equation}
where $\mathbf{f}_i$ are the pose face points, with their corresponding anchor points $\mathbf{a}_{j(i)}$ and anchor normals $\mathbf{n}_{j(i)}$. Both stages optimize only yaw and in-plane translation to preserve the gravity axis. The composite transform $\mathbf{T}_k=\mathbf{T}_2\mathbf{T}_1\mathbf{T}_0$ yields a fused, occlusion-completed canonical point cloud $\bigcup_k\mathbf{T}_k(\mathcal{I}_k)$ for each shape type.

\begin{figure*}[!t]
\centering
\includegraphics[width=1\linewidth]{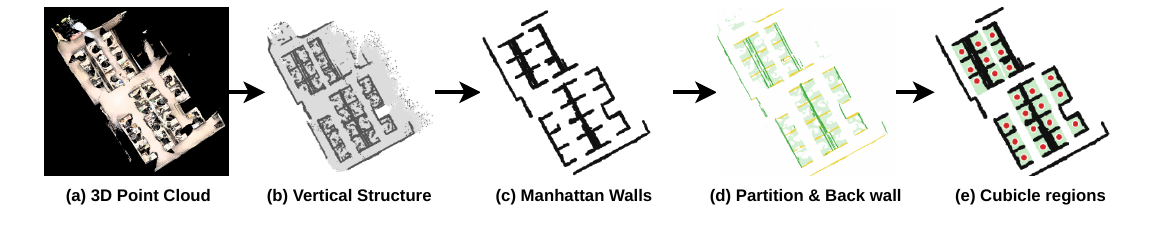}
\caption{Phase 2 topology extraction. (a)~Gravity-aligned cloud (ceiling removed). (b)~Near-vertical points (wall candidates). (c)~Manhattan plane fitting extracts wall skeletons. (d)~Walls classified as partitions (orange) or back walls (blue). (e)~Cutting desk planes with walls isolates $23$ trajectory-independent cubicles (green regions, red centers).}
\label{fig:topology}
\end{figure*}

\subsection{Consensus Denoising and Manhattan Solid Fitting}

The fused cloud is voxelized at resolution $\delta_v$ to enable spatial consensus evaluation. For each voxel $v$, we compute the ratio between the actual observers $N_{act}(v)=\vert{}\{k:\mathcal{I}_k\cap v\neq\emptyset\}\vert{}$ and the expected observers $\sum_k\mathrm{seen}_k(v)$. Cubicle $k$ is expected to see voxel $v$ if it is depth-consistent in a fraction $\rho_0$ of its keyframes $\mathcal{F}_k$:
\begin{equation}
    \begin{aligned} \mathrm{seen}_k(v)&=\mathbb{I}\!\Big(\tfrac{1}{\vert{}\mathcal{F}_k\vert{}}\!\!\sum_{f\in\mathcal{F}_k}\!\mathrm{vis}(\mathbf{T}_k^{-1}v_c,f)\ge\rho_0\Big),\\ \mathrm{vis}(p,f)&=\mathbb{I}\!\big(z_f(p)\le\mathcal{D}_f(\pi_f(p))+\epsilon_0+\kappa\,z_f(p)\big), \end{aligned}
\end{equation}
where $v_c$ is the voxel center, $\mathbf{T}_k^{-1}$ is the inverse canonical transform, and $\mathrm{vis}(p,f)$ evaluated at back-projected point $p$ inside frame $f$ uses camera depth $\mathcal{D}_f$, projection $\pi_f$, camera-frame depth $z_f(p)$, and a distance-adaptive margin governed by $\epsilon_0$ and $\kappa$. Voxels failing a consensus ratio $\tau_c$ are pruned. This multi-view corroboration naturally eliminates pose-drift ghosts supported by a single instance, avoiding the need for arbitrary noise thresholds:
\begin{equation}
    \mathcal{C}(v)=\frac{N_{act}(v)}{\max\!\big(\sum_k\mathrm{seen}_k(v),\,1\big)}\ \ge\ \tau_c.
\end{equation}
\textbf{Manhattan Solid Fitting:} The retained voxels guide structural plane extraction. Axis-aligned planes are extracted using a strict normal gate $\vert{}\mathbf{n}\!\cdot\!\mathbf{a}\vert{}>0.9$ (for axes $\mathbf{a}\in\{\mathbf{e}_x,\mathbf{e}_y,\mathbf{e}_z\}$) and offset-histogram peaks. A plane is retained only if its horizontal footprint occupancy exceeds $\phi_{\min}$. This fill test explicitly discriminates dense architecture from sparse desktop clutter where pure point counts fail. Parallel faces separated within a thickness range $[t_{\min},t_{\max}]$ are paired into volumetric slabs, discretized into axis-aligned boxes, and merged via boolean union into a watertight canonical template $\mathcal{M}_{\text{type}}=\textstyle\bigcup_{b}\mathrm{box}_b$ for each shape type.\textbf{Layout-Consistent Scene Assembly:} Relocating independent templates often causes adjacent instances to interpenetrate, as the true thickness of a shared partition is unobservable from a single cubicle. To resolve this, we globally optimize the partition layout. Let $v_k$ be the position of instance $k$ along the partition axis, and let adjacent instances form chains $\{\mathcal{R}_j\}$. We optimize a global spacing $p$ and per-chain offsets $o_j$:
\begin{equation}
    (p^{\star},\{o_j^{\star}\})=\arg\min_{p,\{o_j\}}\sum_{j}\sum_{k\in\mathcal{R}_j} \big(v_k-(o_j+i_k\,p)\big)^{2},
\end{equation}
where $i_k$ is the integer index of cubicle $k$ within its chain. The shared partition thickness is resolved as $t=p^{\star}-c$, where $c$ is the inner clear span of the canonical template. This constraint guarantees face-to-face board coincidence between neighbours by construction. Each instance is then snapped to the layout grid using its centroid $\mathbf{c}$ to eliminate lever-arm translation: $\mathbf{t}_k'=\mathbf{t}_k+\big(\mathbf{R}(\psi_k)-\mathbf{R}(\hat{\psi}_k)\big)\mathbf{c}$, where $\mathbf{t}_k$ and $\mathbf{R}(\psi_k)$ represent original translation and rotation, and $\mathbf{R}(\hat{\psi}_k)$ is the grid-snapped orientation. Finally, all vertical slabs are extruded to a global floor plane re-estimated via RANSAC on the point band immediately above the floor, recovering the full height omitted during instance segmentation.

\section{Experiments}

We evaluate OfficeTwin on the global subset of Office Hours~\cite{cantu2026office}, a real-world office cubicle benchmark for embodied scene understanding, uniformly sampling $1{,}133$ RGB frames from an open-plan office containing $23$ cubicles ($22$ U-shaped and $1$ L-shaped). The sequences are captured by a mobile robot along fixed corridor paths, resulting in systematic partial-view observations of near-identical cubicles, matching the conditions targeted by our method. The dataset also provides annotated 2D layouts for topology extraction and cubicle decomposition. Since no CAD model is available, Chamfer distance and instance mAP cannot be computed. Instead, we evaluate geometric usability (watertightness, metric scale), redundancy consistency (cross-instance corroboration), and reconstruction success (cubicle recovery and shape-type accuracy). We compare against feed-forward reconstruction (VGGT~\cite{wang2025vggt}, DUSt3R~\cite{wang2024dust3r}), TSDF fusion, and single-image generative models (Hunyuan3D-2~\cite{zhao2025hunyuan3d}, TRELLIS~\cite{xiang2025structured}, SAM3D~\cite{chen2025sam}). \textbf{Video results are available in the supplementary material.}

\subsection{Reconstruction Capability and Watertightness}

Qualitative and quantitative evaluations of the reconstructed geometry are provided in Fig.~\ref{fig:comparison} and Table~\ref{tab:capability}. While feed-forward methods (VGGT, DUSt3R) retain only scale-ambiguous, topology-free point clouds, TSDF fusion yields a metric but severely defective surface. Specifically, TSDF fusion fails on two fronts. Semantically, it lacks object identity: $92.3\%$ of its $2.97$M triangles merge into a single connected component, welding all $23$ cubicles, partitions, desks, and the floor into an inseparable mass that prevents asset re-posing or physical property assignment. Geometrically, it fails to form a closed volume, leaving $260{,}351$ open boundary edges and fragmenting residual triangles into $15{,}197$ detached debris units ($14{,}975$ having fewer than 100 faces), rendering the mesh unusable as a collision proxy. In contrast, OfficeTwin leverages the recovered wall topology to partition the scan into independent workstation instances by classifying structural walls and bounding the desk planes (Fig.~\ref{fig:topology}). It stands as the only approach that simultaneously delivers absolute metric scale, watertight geometry ($\mathbf{0}$ open edges), and object-level decomposition required for high-fidelity physics simulation (Fig.~\ref{fig:comparison}).

\begin{figure*}[!t]
\centering
\includegraphics[width=\textwidth]{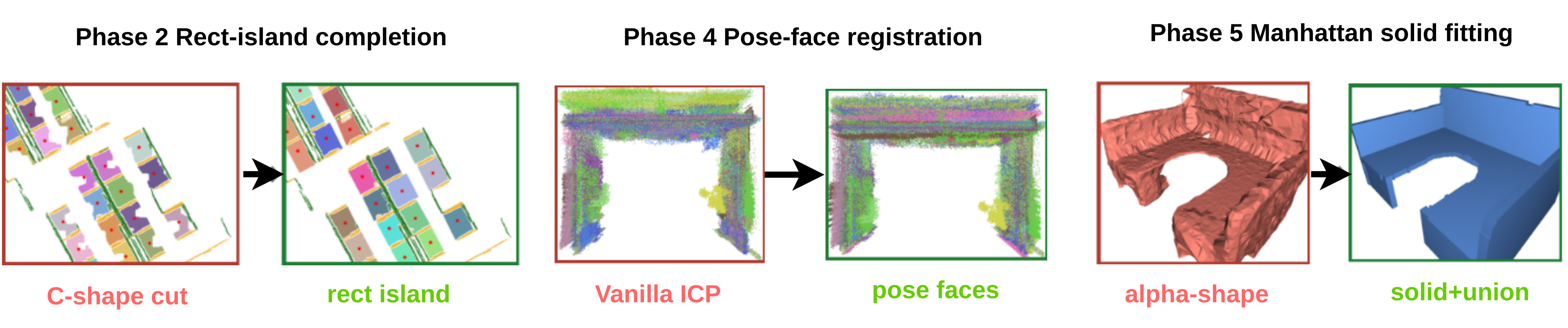}
\caption{Ablation of key designs: failure (left, red) vs full pipeline (right, green). P2: raw C-shaped cut leaks 3D growth, inflating cubicles ($5.80{\times}4.90$\,m vs $1.81{\times}2.42$\,m). P4: vanilla ICP is distracted by cabinet planes (spread $0.137$\,m vs $0.044$\,m on pose faces). P5: alpha-shapes leave open edges ($66$ vs $0$). Removing any step breaks the twin.}
\label{fig:ablation}
\end{figure*}

\begin{table}[!t]
\centering\small
\setlength{\tabcolsep}{4pt}
\begin{tabular}{lcccc}
\toprule
Method & Metric & Watertight & Per-obj. & Sim.\\
 & scale & (open edges) & / inst. & ready\\
\midrule
\multicolumn{5}{l}{\emph{Full-scene reconstruction}}\\
DUSt3R & $\times$ & $\times$~(points) & $\times$ & $\times$\\
VGGT & $\times$ & $\times$~(points) & $\times$ & $\times$\\
TSDF fusion & $\checkmark$ & $\times$~($260$k) & $\times$ & $\times$\\
\textbf{OfficeTwin} & $\checkmark$ & $\checkmark$~($0$) & $\checkmark$ & $\checkmark$\\
\midrule
\multicolumn{5}{l}{\emph{Single cubicle (partial + redundant view)}}\\
VGGT & $\times$ & $\times$ & $\times$ & $\times$\\
DUSt3R & $\times$ & $\times$ & $\times$ & $\times$\\
Hunyuan3D-2 & $\times$ & $\times$ & $\times$ & $\times$\\
TRELLIS & $\times$ & $\times$ & $\times$ & $\times$\\
SAM3D & $\times$ & $\checkmark$~($0$) & $\times$ & $\times$\\
\textbf{OfficeTwin} & $\checkmark$ & $\checkmark$ & $\checkmark$ & $\checkmark$\\
\bottomrule
\end{tabular}
\caption{Reconstruction capability. OfficeTwin alone is metric, watertight, per-object, and simulation-ready; baselines fuse scenes into inseparable surfaces/clouds or fragment instances.}
\label{tab:capability}
\end{table}

\subsection{Partial View and Redundant Structure}
\label{subsec:partial}
To evaluate how sequence-level redundancy resolves unobserved geometry, we reconstruct one cubicle under asymmetric inputs. The input is not held constant across paradigms, because each is defined over a different observation unit. Single-image generative models (Hunyuan3D-2, TRELLIS, SAM3D) accept one view; multi-view feed-forward models (VGGT, DUSt3R) require overlapping views and are given the 21 frames size keyframe window; and OfficeTwin consumes the full sequence, because evidence for occluded surfaces resides in other repeated instances rather than any local window. All baselines are anchored on the same reference frames, so they differ in how many neighbouring views their architecture admits. The experiment does not ask which method performs best under a fixed input budget; it asks what each paradigm produces from its observations, and whether sequence-level redundancy supplies geometry that no local observation contains (Table~\ref{tab:capability}, Fig.~\ref{fig:baselines}). Feed-forward models (VGGT, DUSt3R) only recover geometry within the camera's line-of-sight up to an unknown similarity transform; Poisson surfacing merely introduces spurious sheets rather than partition walls. Generative 3D methods hallucinate geometry without structural priors under partial views. Hunyuan3D-2 leaves occluded regions unreconstructed with surface artifacts, while TRELLIS misproportions the enclosure, widening the footprint relative to the cubicle's true aspect. Although SAM3D yields visually cleaner surfaces, it skews partitions and heights (Fig.~\ref{fig:baselines}, red boxes). Furthermore, spatial redundancy breaks instance identity for the baselines, causing SAM3D to absorb adjacent cubicle parts and erase instance boundaries. Conversely, OfficeTwin uses repeated instances as mutual geometric references, enforcing partition orthogonality and footprint bounds constrained by physical evidence. The structural redundancy that destabilizes baseline segmentations thus enables OfficeTwin to resolve individual workstations into separate, watertight, and metrically scaled solids (Fig.~\ref{fig:baselines}, bottom right).

\subsection{Redundancy-Driven Completion}
We next quantify the efficacy of redundancy-driven completion using repeated cubicles as mutual references (Table~\ref{tab:rce}). While a single partial-view cubicle observes only $21.0\%$ of the fused surface, with an average of $4.6$ supporting instances per surface location, cross-instance fusion recovers missing regions, yielding $68.6\%$ multi-instance corroboration ($\ge 2$ instances). This demonstrates structural completion grounded in physical redundancy rather than unconstrained hallucination. Same-shape instances are aligned tightly into a canonical footprint of $1.85\pm0.06$\,m $\times$ $2.43\pm0.06$\,m with $0/22$ mis-registrations. Constraining the alignment to the outermost pose faces reduces the back-wall plane deviation from $0.137$\,m (vanilla ICP) to $0.044$\,m. The consensus stage subsequently prunes $40.6\%$ of points lacking multi-view support, enabling all $23$ cubicles to form watertight solids. Furthermore, global layout assembly resolves the shared partition thickness that no single cubicle observes, $t=35.4$\,mm, by fitting a global spacing $p^{\star}=2.295$\,m with a $13.7$\,mm residual. Consequently, the adjacent partition faces of all $15$ neighbouring pairs coincide within a median of $0.0$\,mm ($1.1$\,mm worst-case), with zero floor-contact gap.

\begin{table}[t]
\centering
\small
\setlength{\tabcolsep}{4pt}
\begin{tabular}{lc}
\toprule
Metric & Value\\
\midrule
Single-view completeness & $21.0\%$\\
Multi-view corroboration & $68.6\%$\\
Cross-instance support & $4.6/22$\\
Overlay std. ($\sigma$) & $\pm0.06$\,m\\
Registration failures & $0/22$\\
Back-wall spread & $0.137\!\rightarrow\!0.044$\,m\\
Consensus-pruned points & $40.6\%$\\
Watertight instances & $\mathbf{23/23}$\\
Shared-partition thickness & $35.4$\,mm\\
Shared-board coincidence & $0.0$ / $1.1$\,mm (med / max)\\
Floor contact gap & $0.0$\,mm\\
\bottomrule
\end{tabular}
\caption{Redundancy consistency evaluation. Repeated cubicles act as mutual references: cross-instance corroboration lifts surface coverage from $21.0\%$ to $68.6\%$, yielding $23/23$ watertight instances without ground-truth meshes.}
\label{tab:rce}
\end{table}

\subsection{Ablation Study}
We ablate the six design choices on the real scan (Table~\ref{tab:ablation}; the three most consequential cases in Fig.~\ref{fig:ablation}). Removing any component produces a structurally invalid twin, and the failures are structural rather than gradual: mean-density wall filtering drops sparse back walls under clutter, nearest-component assignment cannot decide which cubicle a shared partition belongs to, vanilla ICP locks onto the cabinet planes parallel to the back wall, and alpha-shape meshing has no notion of thickness. Two effects are not visible in the table: without rect-island completion, desk growth leaks through the C-shaped boundary and inflates one U cubicle to $5.80\times4.90$\,m instead of $1.81\times2.42$\,m, and alpha-shape meshing leaves $66$ open edges where Manhattan solid fitting leaves none.

\begin{table}[t]
\centering\small
\setlength{\tabcolsep}{4pt}
\begin{tabular}{llc}
\toprule
Phase & Design choice & Effect\\
\midrule
P2 wall & density mean $\rightarrow$ median & $6\rightarrow22$ cub.\\
P2 assign & per-component $\rightarrow$ per-pixel & $20$U/$2$L $\rightarrow22$U/$1$L\\
P2 detect & traj-carve $\rightarrow$ desk-cut & tuned $\rightarrow$ free\\
P2 region & desk-only $\rightarrow$ rect-island & WT $0\rightarrow23$\\
P4 reg. & ICP $\rightarrow$ pose-faces & $0.137\rightarrow0.044$\,m\\
P5 mesh & alpha-shape $\rightarrow$ solid+union & WT $0/23\rightarrow23/23$\\
\bottomrule
\end{tabular}
\caption{Ablation of design choices per phase on the real scan, demonstrating necessity for correct topology, alignment, and watertightness.}
\label{tab:ablation}
\end{table}

\section{Conclusion}
We presented OfficeTwin, a framework that leverages sequence-level structural redundancy to reconstruct watertight, metrically scaled digital twins of repeated office instances. By registering repeated cubicles into a shared canonical frame and enforcing multi-view consensus on a layout-consistent grid, our method resolves occlusions and instance ambiguities that defeat feed-forward and generative baselines. It recovers all 23 cubicles as watertight solids in 350,s, with future work extending this redundancy-driven paradigm to environments with weaker, non-architectural repetition.

\bibliography{aaai2027}
\end{document}